\documentclass{article}

\usepackage{amsmath}
\usepackage{graphicx}
\usepackage{amsmath}
\usepackage{amssymb}
\usepackage{amsfonts}
\usepackage{algorithmic}
\usepackage{graphicx}
\usepackage{textcomp}
\usepackage{xcolor}
\usepackage{threeparttable}
\usepackage{makecell}
\usepackage{float}
\usepackage{booktabs}
\usepackage{multicol}
\usepackage{multirow}
\usepackage{setspace}
\usepackage{enumitem}
\usepackage{stfloats}
\usepackage{cite}
\usepackage{pifont}
\usepackage{hyperref}
\usepackage{spconf}
\usepackage{makecell}

\title{Learning Audio-Visual embedding for Person Verification in the Wild}
\name{\makecell[c]{Peiwen Sun\textsuperscript{1,2}, Shanshan Zhang\textsuperscript{2}, Zishan Liu\textsuperscript{1}, Yougen Yuan\textsuperscript{2}, \\Taotao Zhang\textsuperscript{2}, Honggang Zhang\textsuperscript{1}, Pengfei Hu\textsuperscript{2}}}
\address{ 
  $^1$School of Artificial Intelligence,\\Beijing University of Posts and Telecommunications, Beijing, China\\
  $^2$ TEG AI, Tencent Inc, Beijing, China}

\graphicspath{{images/}}
\begin{document}
%\ninept
%
\normalsize

\maketitle
% \vspace{-1cm}
\begin{abstract}
\vspace{-0.1cm}

It has already been observed that audio-visual embedding is more robust than uni-modality embedding for person verification.
% A novel mechanism is urgently needed in the audio-visual field to study the utterance level representation from each frame.
Here, we proposed a novel audio-visual strategy that considers aggregators from a fusion perspective.
First, we introduced weight-enhanced attentive statistics pooling for the first time in face verification.
We find that a strong correlation exists between modalities during pooling, so joint attentive pooling is proposed which contains cycle consistency to learn the implicit inter-frame weight.
Finally, each modality is fused with a gated attention mechanism to gain robust audio-visual embedding.
All the proposed models are trained on the VoxCeleb2 dev dataset and the best system obtains 0.18\%, 0.27\%, and 0.49\% EER on three official trial lists of VoxCeleb1 respectively, which is to our knowledge the best-published results for person verification.

\end{abstract}
\begin{keywords}
person verification, audio-visual, cycle consistency
\end{keywords}
\vspace{-0.6cm}
\section{Introduction}
\vspace{-0.3cm}
\label{sec:intro}

% Voice and face are two typical biometric characteristics that have been investigated a lot for the person verification task. Accordingly, the speaker verification and face verification tasks are two separate research topics. 

% The demand of person verification system in life is getting higher and higher. Especially in the wild environment, the robustness of the system is particularly important. The joint verification system based on voice and face has become a universal solution.

Biometrics-based person verification technologies are widely used in access control. In the wild, the speech segments are corrupted with real-world noise including laughter, music, and other sounds. Similarly, face images have variations in pose, image quality, and motion blur. It creates additional challenges for the verification system.

The development of verification in Voxceleb \cite{dataset_1,dataset_2} first appeared in speaker verification. The research \cite{speaker_6} proposed attentive statistics pooling to focus on important frames of speaker verification and get higher discriminative ability than the traditional averaging method. In further research, ECAPA-TDNN \cite{speaker_1} was based on blocks of TDNNs and Squeeze-Excitation(SE) \cite{speaker_7} to reconstruct frame-level features. Meanwhile, The margin-based softmax loss originating from face recognition is widely used for training models\cite{face_2,face_3,face_4,face_5}. These losses were also customized to adapt to the speaker verification task \cite{speaker_8,speaker_13} and achieved SOTA at that time. 
% Many researchers \cite{speaker_9,speaker_10} pay attention to the robustness and explore a lot of related extension work. 
% Recent advances driven by large-scale pre-trained models have taken the task of speaker recognition to a new level. With pre-training, Superb\cite{speaker_11}, Unispeech\cite{speaker_12}, Wavlm\cite{speaker_4}, HuBERT\cite{speaker_5} have achieved excellent performances on multiple sets.

% However, the development of loss is mainly reflected in the face verification. The margin-based softmax loss function is widely used for training face recognition models\cite{face_2,face_3,face_4,face_5}. Without the margin, learned features are not sufficiently discriminative. SphereFace \cite{face_4}, CosFace, \cite{face_5} and ArcFace \cite{face_2} introduced different forms of margin functions. 

The performance of speaker systems would degrade dramatically under wild circumstances \cite{Mult_ref_3}. To solve the problem, researchers have found that simply fusing the scores from speaker embeddings and facial embeddings can yield good results \cite{Mult_ref_1,Mult_ref_2}. For the first time, S. Shon \cite{Mult_ref_3} attempted to fuse audio-visual information with deep learning-based models to achieve better results. As the  transformer has been proposed, it seemed more efficient \cite{Mult_ref_4} to fuse different modalities in large datasets. Previous studies have sought implicit feature expression and supervision patterns from the perspective of network structure.  To further explore the explicit correlations, Y. Liang \cite{Mult_ref_5}  gave person verification a point of view of the HGR maximal correlation.
% In the recent boom of pretrain models, B. Shi \cite{Mult_ref_6} provided an audio-visual identity recognition method based on unsupervised pre-training models.
Most of the research on audio-visual works \cite{Mult_ref_7,Mult_ref_4} still focused on extracting frame-level embeddings and then simply averaging them as an aggregator to get segment-level embeddings.

% In this article, we proposed an audio-visual network that considers aggregators from a fusion perspective. The aggregator that is used to generate a single utterance representation from each frame does not seem to be well explored. 
To further increase aggregator robustness and generalization of utterance level embedding, our network is reconsidered from a fusion perspective and brings a huge leap forward in VoxCeleb. 
As a prerequisite, we enhance the frame weight of the attention statistics pooling\cite{speaker_6} to adjust the situation of the audio-visual training process. Intuitively, facial muscles and syllables of pronunciation have a mutual impact on this task. Then, cycle consistency\cite{cycle_1} was introduced into the aggregator for the inference of the keyframes of two modalities. The method that combines weight-enhanced attentive statistics pooling and cycle consistency is called joint attentive pooling. 
Each frame contributes according to its relevance to the final face representation avoiding equal contributions and preventing accumulated errors.
% And the best system obtains the best-published results for person verification to our knowledge.
Finally, as an analysis, visualization maps are generated to explain how this system interacts between modalities and why it is effective.
% \begin{itemize}[itemsep= 0pt,topsep = -4pt,parsep = 0pt, leftmargin = 2em]
% % \begin{itemize}
% \item Overall effective network and training method on audio-visual verification is proposed and brings a huge leap forward in VoxCeleb1\cite{dataset_1}.
% \item Method based on statistical attention pooling is introduced to the process of unimodality, and better performance are obtained with simpler feature extractors and weaker data.
% \item The tanh activation is introduced to alleviate certain defects in training, and we found that this operation can be beneficial to our network in many ways.
% \item Cycle consistency training is naturally introduced to assist training to explore the prior relationship between facial muscles and pronunciation as known as “complementary rule”.
% \end{itemize}

\vspace{-0.6cm}
\section{Method}
\vspace{-0.3cm}

\begin{figure*}[t]
\centering
\includegraphics[width=16cm]{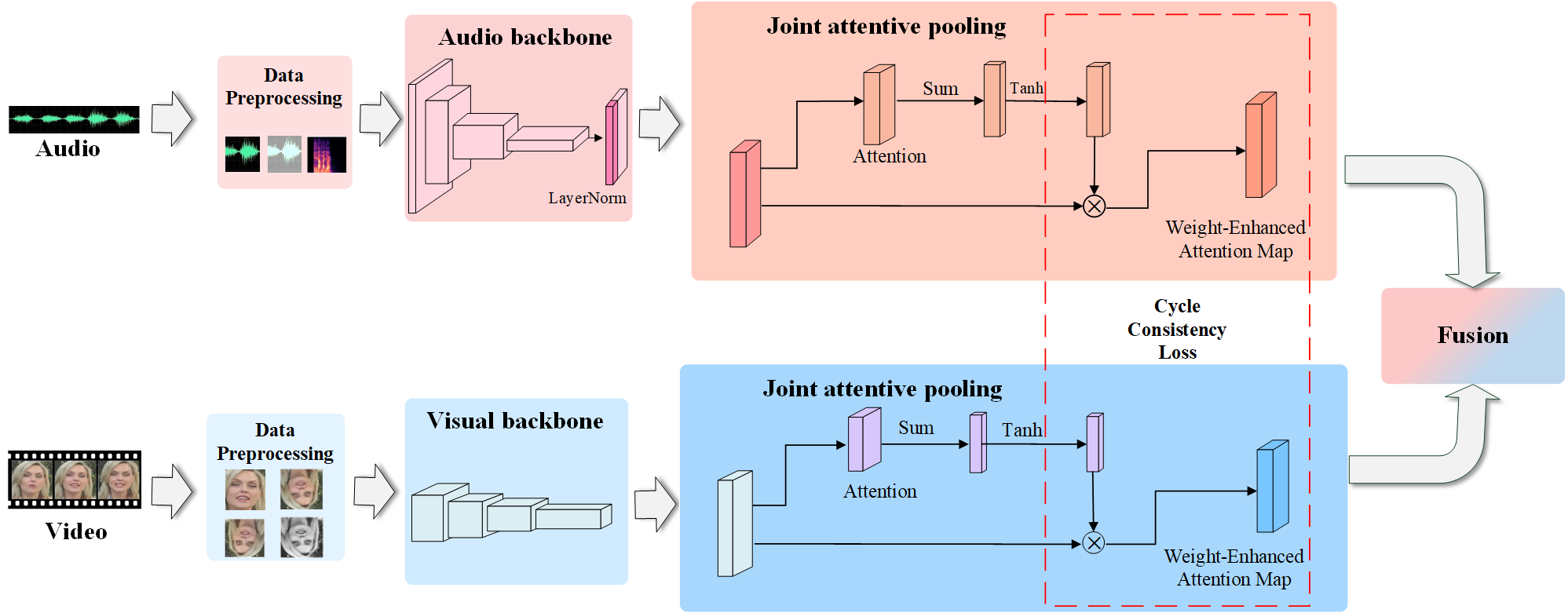}
\vspace{-0.5cm}
\caption{
\small
Overall Network} \label{fig:fusion_net}

\vspace{-0.6cm}  %调整图片与上文的垂直距离

\end{figure*}

% \vspace{-0.3cm}
% \subsection{Feature encoder}
% \vspace{-0.2cm}

The overall network (Fig.\ref{fig:fusion_net}) consists of the backbone of each modality (Sec.\ref{sec:setup}), joint attentive pooling module, and fusion module (Sec.\ref{sec:fusion}). The joint attentive pooling module is the combination of weight-enhanced attentive statistical pooling in Sec.\ref{sec:improve} and cycle consistency learning Sec.\ref{sec:cycle}

\vspace{-0.4cm}
\subsection{Weight-enhanced attentive statistical pooling \label{sec:improve}}
\vspace{-0.2cm}

The data of each modality are pre-processed as stated in Sec.\ref{sec:setup} and sent to the respective encoders. But when visual information is added for joint training, there is a serious drop. To solve the problem, we introduce weight enhanced method of attentive statistical pooling, which is proved to play a positive role in many aspects in the later comparison experiment. The detailed calculation process of weight enhanced method is as follows.

Through the original attentive statistical pooling mechanism \cite{speaker_1}, each frame will be given different weights.

\vspace{-0.3cm}
\begin{equation}
\vspace{-0.01cm}
e^{t, c}=(\boldsymbol{v}^{c})^{T} tanh  \left(\boldsymbol{W} \boldsymbol{h}^{t}+\boldsymbol{b}\right)+k^{c},
\vspace{-0.2cm}
\label{pythagorean}
\end{equation}
%$$
%e^{t, c}=(\boldsymbol{v}^{c})^{T} tanh  %\left(\boldsymbol{W} %\boldsymbol{h}^{t}+\boldsymbol{b}\right)+k^{c}
%$$
% \begin{figure*}[htbp]
% \centering
% \includegraphics[width=18cm]{fusion3.0.png}
% \caption{The training process is divided into two stages. First, two different modalities are trained separately to make their respective weights converge. Then, jointly train the fusion module part for finetune.} \label{fig:fusion_net}
% \end{figure*}
where $h^t$ denotes the activations of the last frame layer at time step $t$. The parameters $\boldsymbol{W} \in \mathbb{R}^{R \times C}$  and $\boldsymbol{b} \in \mathbb{R}^{\check{R \times 1}}$ project the information for self-attention into a smaller $\mathbb{R}$-dimensional representation that is shared across all $C$ channels to reduce the parameter count and risk of overfitting.

This information is transformed to a channel-dependent self attention score through a linear layer with weights $\boldsymbol{v}^{c} \in \mathbb{R}^{R \times 1}$ and bias $k^c$.

However, here comes a problem. It is found that the value of temporal attention $\lambda_t$ has a comparatively small standard deviation. That is to say, the keyframes in the time domain are not obvious. To make the attention more bifurcated and to help the learning of the subsequent layers, we enhance the weight of each frame by the following equation.
%However, when we try to maintain time dimension retained and add or eliminate other dimensions, it can be found that temperal attention $\lambda_t$ has relatively small standard deviation, which means $\lambda_t$ is more concentrated than expected.
% \vspace{-0.1cm}
% \begin{equation}
% \vspace{-0.1cm}
% \lambda^t = sum_{temporal}(e^{t, c}) \label{pythagorean}
% \end{equation}

%$$
%\lambda^t = sum_{temporal}(e^{t, c})
%$$

\vspace{-0.25cm}
\begin{equation}
   e^{t, c}_{tanh} = e^{t, c}*\frac{\lambda_{tanh}^t}{\lambda^t},
\vspace{-0.1cm}
\end{equation}
while the attention map $e^{t, c} \in \mathbb{R}^{T \times A}$ and the projected attention map $e^{t, c}_{tanh} \in \mathbb{R}^{T \times A}$ is given by
\vspace{-0.25cm}
\begin{equation}
    \lambda^t = sum_{temporal}(e^{t, c})\label{pythagorean1},
\vspace{-0.2cm}
\end{equation}
\begin{equation}
   \lambda_{tanh}^t = mu(\lambda^t)*tanh(\frac{\lambda^t-mu(\lambda^t)}{std(\lambda^t)}) + mu(\lambda^t).\label{pythagorean2}
\vspace{-0.15cm}
\end{equation}
% \vspace{-0.5cm}
$mu(\cdot)$, $std(\cdot)$ is mean and standard deviation of the matrix, $\lambda^t \in \mathbb{R}^{T \times 1}$ denotes the temporal attention; 
% $e^{t, c} \in \mathbb{R}^{T \times A}$ denotes the attention map that obtain from last step; $e^{t, c}_{tanh} \in \mathbb{R}^{T \times A}$ denotes the attention map projected from $e^{t, c}$. 
The activated weights are projected into a more dispersed and polarized space $\mathbb{R}^{T \times 1}$ and preserve the mean. If not, fine-grained granularity brings the difficulties of convergence for further experiments. The activation method can avoid non-convergence, alleviate overfitting and be beneficial to the later learning of audio-visual weight interactions.

So compared to the original attention map, it is using the new map $e^{t, c}_{tanh}$ instead of the original map $e^{t, c}$. This scalar score $e^{t, c}_{tanh}$ then normalized over all frames by applying the softmax function channel-wise across time: 
\vspace{-0.25cm}
\begin{equation}
\vspace{-0.1cm}
  \alpha^{t, c}=\frac{\exp \left(e^{t, c}_{tanh}\right)}{\sum_{\tau}^{T} \exp \left(e^{t, c}_{tanh}\right)} .
\vspace{-0.15cm}
\end{equation}

The self-attention score $\alpha_{t, c}$ represents the importance of each frame given the channel and is used to calculate the weighted statistics of channel $c$. Then the weighted mean vector and weighted standard deviation vector can be calculated as
\vspace{-0.3cm}
\begin{equation}
\vspace{-0.1cm}
    \tilde{\mu}^{c}=\sum_{t}^{T} \alpha^{t, c} h^{t, c} ,
\end{equation}
\vspace{-0.13cm}
\begin{equation}
    \tilde{\sigma}^{c}=\sqrt{\sum_{t}^{T} \alpha^{t, c} (h^{t, c})^{2}-(\tilde{\mu}^{c})^{2}} .
\end{equation}

% \vspace{-0.6cm}
% \subsection{Activation analysis\label{sec:activation_analysis}}
% \vspace{-0.2cm}
% The motivation for activation can be actually demonstrated in 3 aspects. Firstly, it is easy to know from the observation experiment that the temporal attention variance $std(\lambda^t)$ of the frame is small, that is, there is no significant difference in temporal domain. Secondly, if we train both of the modality in for joint training, there will be a serious drop in visual for number of visual frames is much fewer than audio. Finally, if these weights are directly used for interactive learning in Fig.\ref{fig:cycle}, it is easy to produce the problem of under fitting or over fitting due to the fine granularity. So we hope to increase the difference between frames, and make a comparative experiment as follows.

% \vspace{-0.3cm}
% \section{Audio-visual learning network}
% \vspace{-0.1cm}
\vspace{-0.6cm}
\subsection{Cycle consistency learning\label{sec:cycle}}
\vspace{-0.2cm}

From the perspective of face verification, researchers such as \cite{face_1} hope to eliminate the impact of facial expressions on face verification. 
Meanwhile, from the perspective of speaker verification, some speech information of audio interval is not helpful for speaker verification.
Then there is a natural idea, hoping to establish some potential probability relationships of keyframes between the face verification and speaker verification, which is called the “complementary rule”.

Essentially, we want to build a network where keyframes between different modalities can be derived from each other. And naturally, there comes a loop. The network used to construct the supervised loop is two encoders consisting of 3-layer mlp to encode the weights of simple temporal frames.

\begin{figure}[htbp]
\centering
\includegraphics[width=8.5cm]{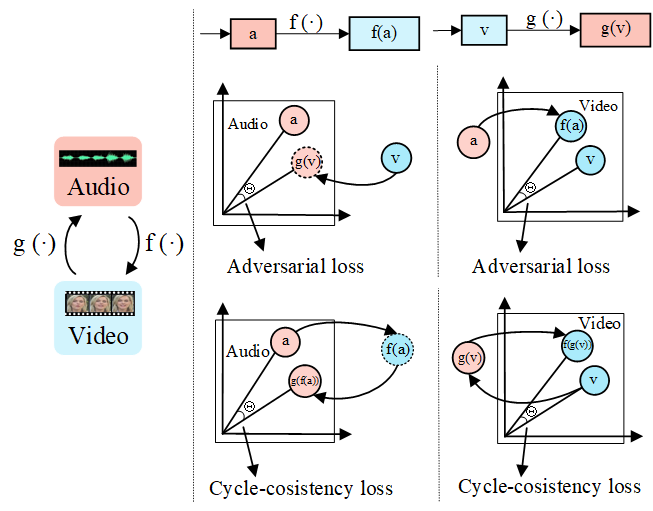}
\vspace{-0.4cm}
\caption{
\small
Proposed adversarial loss and cycle consistency loss} \label{fig:cycle}
\vspace{-0.5cm}  %调整图片与上文的垂直距离
\end{figure}

Here we generalize adversarial loss and cycle consistency loss into regression tasks (Fig.\ref{fig:cycle}). Red and blue blocks represent the temporal attention belonging to different modes respectively, while the solid and dashed lines represent the temporal attention under real conditions and the predicted temporal attention respectively.
The adversarial loss follows the previous paper\cite{cycle_1} to serve a similar purpose, but it is not really adversarial. It ensures each encoder can evolve smoothly, which means the encoder can produce more realistic temporal weights.
The cycle consistency loss ensures that temporal weights from different sources in the same modality have the same intensity.

Adversarial loss and cycle consistency loss only involve the mapping of temporal attention $\lambda_{tanh}^t$, and we express the objective as:
\vspace{-0.4cm}
\begin{equation}
\vspace{-0.35cm}
\begin{split}
    &\mathcal{L}_{\text {adv}}=2 - <a,g(v)> -<v,f(a)>,
    \end{split}
\end{equation}
\vspace{-0.5cm}
\begin{equation}
\begin{split}
    &\mathcal{L}_{\text {cycle}}=2 - <v,f(g(v))> -<a,g(f(a))>,
        \end{split}
\end{equation}
\vspace{-0.7cm}
\begin{equation}
 a = \lambda_{tanh}^{t,a}, v = \lambda_{tanh}^{t,v},
\vspace{-0.2cm}
\end{equation}
while $g(\cdot)$ and $f(\cdot)$ represent 2 MLP encoders separately, ${\mathcal{L}_\text {adv}}$ and ${\mathcal{L}_{\text {cycle}}}$ denote adversarial loss and cycle consistency loss, $<\cdot\ ,\ \cdot>$ is the cosine similarity operator.

Each direction of the loop generates 2 losses which force the attentive statistics pooling to learn potential information. 

Therefore, the overall loss can be calculated as
\vspace{-0.3cm}
\begin{equation}
    \mathcal{L}=\mathcal{L}_{\text {AAM}}+\beta\mathcal{L}_{\text {adv}}+\gamma\mathcal{L}_{\text {cycle}},
\vspace{-0.3cm}
\end{equation}
while $\mathcal{L}_{\text {AAM}}$ is AAM softmax loss, $\beta$ and $\gamma$ are constants that are used to control and adjust the weights of different losses. And their values can be determined by training experience.

% Compared with the L2 loss, the supervision of cosine similarity is weakened. Cosine similarity is more concerned with the direction of temporal attention, that is, more concerned with the relative relationship between the weights of different time frames of each modality.

Compared with the L2 loss, cosine distance is more concerned with the angular relationship between vectors than their absolute size. Using cosine distance as the loss can make the inferred position of the keyframe more accurate. In this case, the unnormalized cosine distance is often better than the Euclidean distance to give the optimal solution.

\vspace{-0.4cm}
\subsection{Fusion strategy\label{sec:fusion}}
\vspace{-0.2cm}
% There have been many explorations in fusion methods in recent years. Bilinear pooling \cite{fusion_1}, transformers\cite{fusion_2} are used to extract mutual features  that have time-series input relationships or logical correspondences. 
% After obtaining the utterance level representation of each modality, it is time for the fusion block to come into play. 
% In terms of audio-visual tasks, there is no necessary relationship between the utterance level features of the face and the speaker after pooling.

Simple Soft Attention Fusion \cite{fusion_2} is implemented. But we use the fully connected layer instead of the transformer layer in the original model to achieve the best fusion effect of the model.

For the joint embedding generated by the network, the convergence can be accelerated by some tricks during the training process. We follow \cite{Mult_ref_8} to impose an orthogonality constraint on the fused embeddings. 
% Although further experiments have shown that adding orthogonal constraints has little effect on the final results. 
We hold the point of view that it is mainly used to accelerate convergence.

% \begin{figure}[t]
% \centering
% \includegraphics[width=9cm]{fusion_block2.0.png}
% \vspace{-1cm}
% \caption{
% \footnotesize
% Fusion block} \label{fig:fusion_block}
% \vspace{-0.5cm}  %调整图片与上文的垂直距离
% \end{figure}

\vspace{-0.4cm}
\section{Expriment Setup\label{sec:setup}}
\label{sec:setup}
\vspace{-0.3cm}

VoxCeleb1\&2 \cite{dataset_1,dataset_2} are used in our experiment. VoxCeleb is an audio-visual dataset consisting of 2,000+ hours of short human speech clips extracted from interview videos on YouTube. For model training, we use the development set of VoxCeleb2, which contains 5,994 speakers. To better evaluate the performance of our network, we adopt 3 trials in VoxCeleb1. For audio data, 80-dimensional Fbank features are extracted with a 25 ms window and 10 ms frameshift, and augmentation with the random mask is added along both the time and frequency domain. Then we do the cepstral mean on the Fbank features. The MUSAN \cite{dataset_5} and RIR Noise datasets \cite{dataset_6} are used as noise sources and room impulse response functions, respectively. For each video segment, we extracted 6 fps in VoxCeleb1 \cite{dataset_1} and 25 fps in VoxCeleb2 \cite{dataset_2} datasets. Then use the similarity transformation to map the face region to the same shape (1 × 128 × 128), which means that we use grey images instead of RGB. Finally, we normalize each image's pixel value to reside in the range of [-0.5, 0.5]. The advanced face detection methods \cite{other_1,other_2} and datasets \cite{dataset_3,dataset_4} are wilder and more fine-grained. So face detection and face alignment are not employed during pre-processing since rough face detection has been performed in VoxCeleb datasets. \cite{dataset_1, dataset_2}.

For the audio encoder, we use ECAPA-TDNN \cite{speaker_1} into which Fbank feature is fed to extract speaker embedding. We made only a few changes to ECAPA-TDNN, which adds additional scale information to the second layer. For the visual encoder, the face feature is extracted by the IResNet18 backbone same as \cite{face_2}.

% Same as the implementation in \cite{speaker_1}, we also use the additive angular margin (AAM) \cite{face_2} loss in the training process for model optimization.

% At the training level, the training of this fusion network is finetuned on the single-modal network training.When we train the overall network, we tend to train the fusion modules of 2 epochs first and then update the overall network.

The training process is divided into two stages. Two different modalities are trained separately first and then finetune in the overall network. All models are trained using AAM softmax loss with a margin of 0.5 and a scaling factor of 30 with 32 NVIDIA Tesla P40s. We use the SGD optimizer with an initial learning rate of 0.01 and decrease the learning rate by 50\% every 2 epochs. As an experimental result, $\beta$ and $\gamma$ are set to 1 and 0.5. Weight decay is set to 5e-4 to avoid overfitting and the global batch size is 320.
 
We use cosine distance with adaptive s-norm\cite{speaker_14} for scoring. Then we report the Equal Error Rate (EER) and minimum Detection Cost Function (minDCF) with $P_{target} = 0.01$ and $C_{FA} = C_{Miss} = 1$ for performance evaluation.

\begin{figure}[tp]
\centering
\includegraphics[width=9cm]{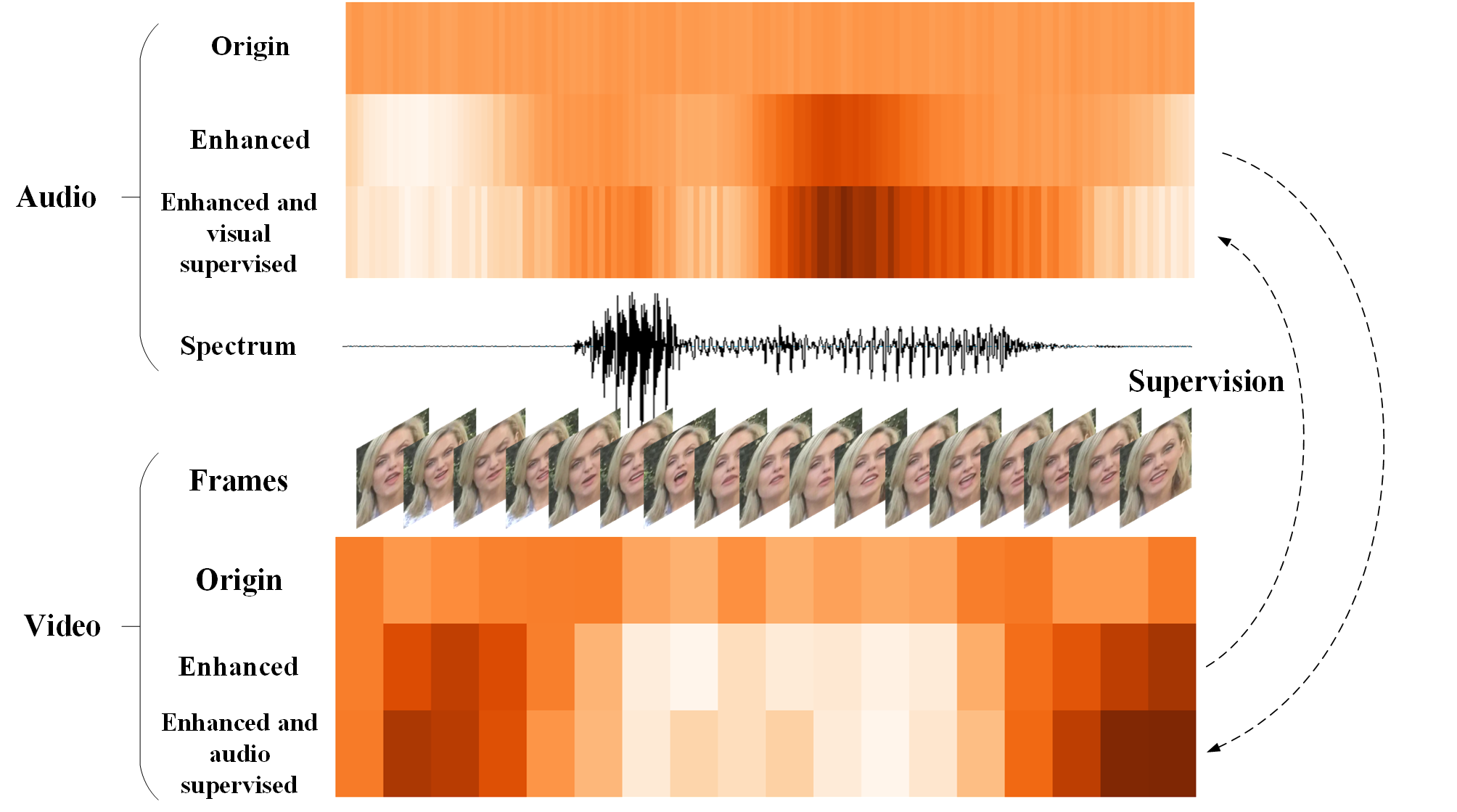}
\vspace{-0.7cm}
\caption{
\small
Attention Heatmap of a 0.8-sec video clip} \label{fig:heatmap}
\vspace{-0.6cm}
\end{figure}
%\label{sec:typestyle}
\vspace{-0.3cm}

\vspace{-0.2cm}
\section{Result Analysis}
\vspace{-0.3cm}

\begin{table*}[b]
\vspace{-0.5cm}
\centering
\setlength{\tabcolsep}{1.5mm}
\caption{\small
Performance of proposed network}
\label{tab:1}
%\resizebox{16cm}{45mm}
\scalebox{0.8}{ 
\begin{threeparttable}
    \begin{tabular}{*{9}{c}}
      \toprule
      \multicolumn{1}{c}{\multirow{2}{*}{\textbf{Type}}} &
      \multicolumn{1}{c}{\multirow{2}{*}{\textbf{Architecture}}}&
      \multicolumn{1}{c}{\multirow{2}{*}{\textbf{\#Modality}}} &
      \multicolumn{2}{c}{\textbf{VoxCeleb1-O}} &
      \multicolumn{2}{c}{\textbf{VoxCeleb1-E}} &
      \multicolumn{2}{c}{\textbf{VoxCeleb1-H}} \\
      \cmidrule(lr){4-5}\cmidrule(lr){6-7}\cmidrule(lr){8-9}
      & & &{\textbf{EER(\%)}} & {\textbf{MinDCF}} & {\textbf{EER(\%)}} & {\textbf{MinDCF}}& {\textbf{EER(\%)}} & {\textbf{MinDCF}}\\
      \midrule
      \multirow{2}*{Unimodal}&
      ECAPA-TDNN\cite{speaker_1} & A & 0.87 &  0.107 & 1.12 & 0.132 & 2.12  &0.210\\
        ~ &SimAM-Resnet\cite{speaker_2} & A & 0.64 & 0.067  & 0.84 & 0.089 & 1.49  &0.146\\
      \midrule
      \multirow{14}*{Audio-visual}&
          \multirow{5}*{Z. Chen\cite{Mult_ref_7}} 
          ~ & \ding{172}A & 2.31 & - & 2.23 & - & 3.78 & -\\
          ~& ~ & \ding{173}V & 2.26 & - & 1.54 & - & 2.37 & -\\
            ~ & ~ & \ding{174}fusion & 0.59 & - & 0.43 & - & 0.74 & -\\
            ~ & ~ & ensemble(\ding{172},\ding{173}) & 0.51 & - & 0.43 & - & 0.78 & -\\
            ~ & ~ & ensemble(\ding{172},\ding{173},\ding{174}) & 0.50 & - & 0.38 & - & 0.68 & -\\
    \cmidrule(lr){2-9}
          &\multirow{3}*{Y. Qian\cite{Mult_ref_4}} 
          ~ & A & 1.62 & - & 1.75 & - & 3.16 & -\\
          ~& ~& V & 3.04 & - & 2.18 & - & 4.23 & -\\
            ~&~ & fusion & 0.71 & - & 0.48 & - & 0.85 & -\\
        \cmidrule(lr){2-9}
     & \multirow{6}*{Ours} 
     ~ & A w/o enhanced$\dagger$ & 0.98 & 0.140 & 1.24 & 0.163 & 2.30 & 0.264\\
     ~ & ~ & \ding{175}A & 0.99 & 0.140 & 1.24 & 0.163 & 2.27 & 0.264\\
      ~& ~ & V w/o enhanced$\dagger$ & 2.59 & 0.214 & 1.88 & 0.198 & 3.39 & 0.297\\
      ~& ~ & \ding{176}V & 1.44 & 0.147 & 1.28 & 0.157 & 2.14 & 0.230\\
    %   ~&Ours(w/o)\dagger & 2.59 & 0.214 & 1.88 & 0.198 & 3.39 & 0.297\\
        ~&~ & fusion w/o cons$\dagger$ & 0.22 & 0.022 & 0.27 & 0.035 & 0.52 & 0.058\\
        ~&~ & \ding{177}fusion & 0.18 & 0.017 & 0.26 & 0.035 & 0.49 & 0.057\\
      \midrule
        Audio-visual & Ours & ensemble(\ding{175},\ding{176},\ding{177}) & \textbf{0.14} & \textbf{0.012} & \textbf{0.21} & \textbf{0.028} & \textbf{0.37} & \textbf{0.046}\\
      \bottomrule
    \end{tabular}
    \begin{tablenotes}
    	\normalsize
    % 	\item[$\dagger$]Since 4 samples of the audio of voxceleb1 do not have corresponding images, the trials of these samples are eliminated in the visual and fusion task.
    % \vspace{-0.3cm}
    \item[$\dagger$]``w/o cons" means ``without cycle consistency" and ``w/o enhanced" means ``without weight enhanced method".
    \end{tablenotes}
\end{threeparttable}
}
\vspace{0cm}
\end{table*}

From table \ref{tab:1}, we observe that the weight-enhanced method for unimodality will not bring obvious changes to speech while using activation in the face alone can be a relief for the generalization pressure of attentive statistic pooling. 
% The reason for the phenomenon comes from the fact that faces have fewer frames and fewer variations.
And the later experiments show that the activation operation on a single modality brings better performance on the joint audio-visual model. In terms of vision alone, compared with the results of works \cite{Mult_ref_4,Mult_ref_7}, our proposed method uses less image information (grey image instead of RGB), less data preparation (without face detection and face alignment), a shallower extractor (IResNet18 instead of ResNet34\cite{Mult_ref_4} or SE-ResNet50\cite{Mult_ref_7}). This method enables us to gain a competitive reduction of EER to 63.7\% of the previous baseline of vision by adopting a higher sampling rate.

\begin{figure}[t]
\centering
\includegraphics[width=7.5cm,height=5.3cm]{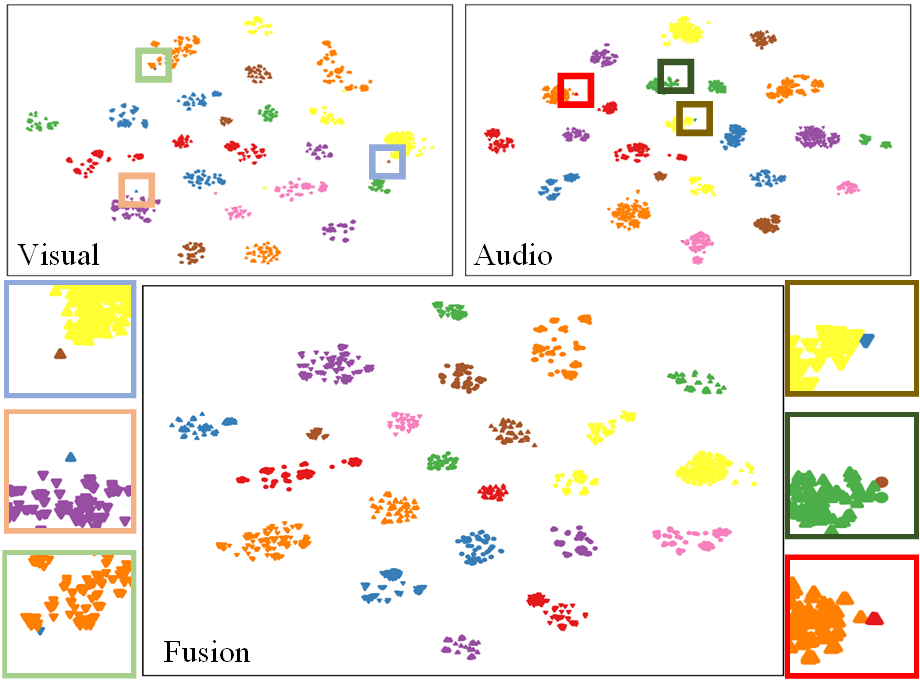}
\vspace{-0.5cm}
\caption{
\small
Sample visualization} \label{fig:tsne}
\vspace{-0.7cm}  %调整图片与上文的垂直距离
\end{figure}

%\label{sec:typestyle}

The cycle-consistency loss will finally converge to around 0.03-0.05. As the loss converges and each encoder is analyzed separately, the transformation from vision to audio is easier to converge than the transformation from audio to vision. This shows that the ``complementary rule" is mainly reflected in ``vision-supervised audio". From the results in Table \ref{tab:1}, the use of cycle loss indeed enhances the weight of important time frames. 
% Compared with related work on face recognition under the same dataset, we use shallower feature extractors and less image information, and only use time series information to significantly improve the performance.
Compared to audio-visual person verification, the fusion network alone can greatly reduce the EER down to 30.7\% of the previous baseline. The simple score ensemble of multiple systems is still valid for this experiment. The best results are achieved after a simple score ensemble.

Attention map Fig.\ref{fig:heatmap} of the output of the two different layers is generated to verify the ``complementary rule". We can intuitively see that the time frame weight is adjusted adaptively and the two modalities are roughly complementary.
The most accurate time frame for face verification is often accompanied by rare facial muscle movements; while the most accurate time frame for speaker verification is often accompanied by a wide range of facial muscle movements.
% And compared with the SOTA results of speaker verification based on pretraining or non-pretraining, it has great advantages.

About 20 hard samples were selected from the VoxCeleb1-H dataset for t-SNE dimensionality reduction visualization analysis. It can be seen from Fig.\ref{fig:tsne} that each diagram of a single modality can easily find 5-6 outliers. But our network is more robust to these outliers. This phenomenon well demonstrates the robustness of the proposed network.

\vspace{-0.5cm}
\section{Conclusion\label{sec:Conclusion}}
\label{sec:Conclusion}
\vspace{-0.3cm}

Generally, we proposed a novel audio-visual strategy that considers aggregators from a fusion perspective. Compared with previous aggregators, we proposed joint attentive pooling based on cycle consistency as a generic aggregator for the first time. 
It significantly reduces the EER down to 28.0\% of the popular systems.
And analysis shows the robustness of our network and the existence of “complementary rules”. In future work, the audio-visual unsupervised pretrain model will be explored to reach a higher level.

% \vspace{-0.5cm}
% \section{Acknowledgment}
% \vspace{-0.25cm}
% Experiments have been carried out during the internship of Peiwen Sun in Tencent Inc.
\vspace{-0.4cm}

\begin{spacing}{1.5}

\setlength{\itemsep}{-2mm}

\bibliographystyle{ieeetr} 

\setstretch{1}

%\bibliography{strings,refs}
\footnotesize

\bibliography{ref}

\begin{thebibliography}{10}

\bibitem{dataset_1}
A.~Nagrani, J.~S. Chung, W.~Xie, and A.~Zisserman, ``Voxceleb: Large-scale
  speaker verification in the wild,'' {\em Computer Science and Language},
  2019.

\bibitem{dataset_2}
J.~S. Chung, A.~Nagrani, and A.~Zisserman, ``Voxceleb2: Deep speaker
  recognition,'' in {\em INTERSPEECH}, 2018.

\bibitem{speaker_6}
K.~Okabe, T.~Koshinaka, and K.~Shinoda, ``Attentive statistics pooling for deep
  speaker embedding,'' {\em arXiv preprint arXiv:1803.10963}, 2018.

\bibitem{speaker_1}
B.~Desplanques, J.~Thienpondt, and K.~Demuynck, ``Ecapa-tdnn: Emphasized
  channel attention, propagation and aggregation in tdnn based speaker
  verification,'' in {\em Interspeech}, pp.~3830--3834, 2020.

\bibitem{speaker_7}
J.~Hu, L.~Shen, and G.~Sun, ``Squeeze-and-excitation networks,'' in {\em Proc.
  CVPR}, pp.~7132--7141, 2018.

\bibitem{face_2}
J.~Deng, J.~Guo, N.~Xue, and S.~Zafeiriou, ``Arcface: Additive angular margin
  loss for deep face recognition,'' in {\em Proc. CVPR}, June 2019.

\bibitem{face_3}
Y.~Huang, Y.~Wang, Y.~Tai, X.~Liu, P.~Shen, S.~Li, J.~Li, and F.~Huang,
  ``Curricularface: adaptive curriculum learning loss for deep face
  recognition,'' in {\em Proc. CVPR}, pp.~5901--5910, 2020.

\bibitem{face_4}
W.~Liu, Y.~Wen, Z.~Yu, M.~Li, B.~Raj, and L.~Song, ``Sphereface: Deep
  hypersphere embedding for face recognition,'' in {\em Proc. CVPR},
  pp.~212--220, 2017.

\bibitem{face_5}
H.~Wang, Y.~Wang, Z.~Zhou, X.~Ji, D.~Gong, J.~Zhou, Z.~Li, and W.~Liu,
  ``Cosface: Large margin cosine loss for deep face recognition,'' in {\em
  Proc. CVPR}, pp.~5265--5274, 2018.

\bibitem{speaker_8}
L.~Wan, Q.~Wang, A.~Papir, and I.~L. Moreno, ``Generalized end-to-end loss for
  speaker verification,'' in {\em Proc. ICASSP}, pp.~4879--4883, IEEE, 2018.

\bibitem{speaker_13}
X.~Xiang, S.~Wang, H.~Huang, Y.~Qian, and K.~Yu, ``Margin matters: Towards more
  discriminative deep neural network embeddings for speaker recognition,'' in
  {\em APSIPA ASC}, pp.~1652--1656, IEEE, 2019.

\bibitem{Mult_ref_3}
S.~Shon, T.-H. Oh, and J.~Glass, ``Noise-tolerant audio-visual online person
  verification using an attention-based neural network fusion,'' in {\em Proc.
  ICASSP}, pp.~3995--3999, IEEE, 2019.

\bibitem{Mult_ref_1}
S.~O. Sadjadi, C.~S. Greenberg, E.~Singer, D.~A. Reynolds, L.~P. Mason,
  J.~Hernandez-Cordero, {\em et~al.}, ``The 2019 nist audio-visual speaker
  recognition evaluation.,'' in {\em Odyssey}, pp.~259--265, 2020.

\bibitem{Mult_ref_2}
J.~Alam, G.~Boulianne, L.~Burget, M.~Dahmane, M.~D. S{\'a}nchez,
  A.~Lozano-Diez, O.~Glembek, P.-L. St-Charles, M.~Lalonde, P.~Matejka, {\em
  et~al.}, ``Analysis of abc submission to nist sre 2019 cmn and vast
  challenge.,'' in {\em Odyssey}, pp.~289--295, 2020.

\bibitem{Mult_ref_4}
Y.~Qian, Z.~Chen, and S.~Wang, ``Audio-visual deep neural network for robust
  person verification,'' {\em IEEE. Trans Audio Speech Lang Process}, vol.~29,
  pp.~1079--1092, 2021.

\bibitem{Mult_ref_5}
Y.~Liang, F.~Ma, Y.~Li, and S.-L. Huang, ``Person recognition with hgr maximal
  correlation on multimodal data,'' in {\em Proc. ICPR}, pp.~1--8, IEEE, 2021.

\bibitem{Mult_ref_7}
Z.~Chen, S.~Wang, and Y.~Qian, ``Multi-modality matters: A performance leap on
  voxceleb.,'' in {\em INTERSPEECH}, pp.~2252--2256, 2020.

\bibitem{cycle_1}
J.-Y. Zhu, T.~Park, P.~Isola, and A.~A. Efros, ``Unpaired image-to-image
  translation using cycle-consistent adversarial networks,'' in {\em Proc.
  CVPR}, pp.~2223--2232, 2017.

\bibitem{face_1}
J.~Chang, Z.~Lan, C.~Cheng, and Y.~Wei, ``Data uncertainty learning in face
  recognition,'' in {\em Proc. CVPR}, pp.~5710--5719, 2020.

\bibitem{fusion_2}
Y.-H.~H. Tsai, S.~Bai, P.~P. Liang, J.~Z. Kolter, L.-P. Morency, and
  R.~Salakhutdinov, ``Multimodal transformer for unaligned multimodal language
  sequences,'' in {\em Proc. ACL}, vol.~2019, p.~6558, NIH Public Access, 2019.

\bibitem{Mult_ref_8}
M.~S. Saeed, M.~H. Khan, S.~Nawaz, M.~H. Yousaf, and A.~Del~Bue, ``Fusion and
  orthogonal projection for improved face-voice association,'' in {\em Proc.
  ICASSP}, pp.~7057--7061, 2022.

\bibitem{dataset_5}
D.~Snyder, G.~Chen, and D.~Povey, ``Musan: A music, speech, and noise corpus,''
  {\em arXiv preprint arXiv:1510.08484}, 2015.

\bibitem{dataset_6}
T.~Ko, V.~Peddinti, D.~Povey, M.~L. Seltzer, and S.~Khudanpur, ``A study on
  data augmentation of reverberant speech for robust speech recognition,'' in
  {\em Proc. ICASSP}, pp.~5220--5224, IEEE, 2017.

\bibitem{other_1}
J.~Deng, J.~Guo, E.~Ververas, I.~Kotsia, and S.~Zafeiriou, ``Retinaface:
  Single-shot multi-level face localisation in the wild,'' in {\em Proc. CVPR},
  2020.

\bibitem{other_2}
J.~Guo, J.~Deng, A.~Lattas, and S.~Zafeiriou, ``Sample and computation
  redistribution for efficient face detection,'' in {\em Proc. ICLR}, 2021.

\bibitem{dataset_3}
S.~Yang, P.~Luo, C.~C. Loy, and X.~Tang, ``Wider face: A face detection
  benchmark,'' in {\em Proc. CVPR}, 2016.

\bibitem{dataset_4}
X.~Zhu, Z.~Lei, X.~Liu, H.~Shi, and S.~Z. Li, ``Face alignment across large
  poses: A 3d solution,'' in {\em Proc. CVPR}, June 2016.

\bibitem{speaker_14}
P.~Matejka, O.~Novotn{\`y}, O.~Plchot, L.~Burget, M.~D. S{\'a}nchez, and
  J.~Cernock{\`y}, ``Analysis of score normalization in multilingual speaker
  recognition.,'' in {\em Interspeech}, pp.~1567--1571, 2017.

\bibitem{speaker_2}
X.~Qin, N.~Li, C.~Weng, D.~Su, and M.~Li, ``Simple attention module based
  speaker verification with iterative noisy label detection,'' in {\em Proc.
  ICASSP}, pp.~6722--6726, IEEE, 2022.

\end{thebibliography}
\end{spacing}
\vspace{10pt}
\end{document}